\documentclass[sigconf, nonacm,screen]{acmart}

\newcommand{\anonevolve}{AlphaEvolve}
\newcommand{\alphaarchitect}{ArchAgent}
\newcommand{\policyone}{\texttt{Policy31}}
\newcommand{\policyonept}{\texttt{Policy31-Tuned}}

\newcommand{\policythree}{\texttt{Policy61}}
\newcommand{\policyfour}{\texttt{Policy62}}

\usepackage{algorithmic}
\usepackage{graphicx}
\usepackage{textcomp}
\usepackage{xcolor}
\usepackage{listings}
\usepackage{booktabs}
\usepackage{multirow}
\usepackage{multicol}
\usepackage{adjustbox}
\usepackage{siunitx}

\usepackage{makecell}
\usepackage{multirow}

\lstdefinestyle{cppstyle}{
    language=C++,
    backgroundcolor=\color{black!5},
    commentstyle=\color{green!60!black},
    keywordstyle=\color{blue},
    numberstyle=\tiny\color{gray},
    stringstyle=\color{purple},
    basicstyle=\ttfamily\footnotesize,
    breakatwhitespace=false,
    breaklines=true,
    captionpos=b,
    keepspaces=true,
    numbers=none,
    numbersep=5pt,
    showspaces=false,
    showstringspaces=false,
    showtabs=false,
    tabsize=2
}
\lstdefinelanguage{codediff}{
  morecomment=[f][\color{red!80!black}]{-}, %
  morecomment=[f][\color{green!60!black}]{+}, %
  basicstyle=\ttfamily,
  commentstyle=\color{gray},
}

\setcopyright{none} %

\begin{document}

\title{{\alphaarchitect}: Agentic AI-driven Computer Architecture Discovery}

\author{
  Raghav Gupta\textsuperscript{1$\dagger$}, 
  Akanksha Jain\textsuperscript{2}, 
  Abraham Gonzalez\textsuperscript{2}, 
  Alexander Novikov\textsuperscript{3}, 
  Po-Sen Huang\textsuperscript{3}, \\[0.5ex] %
  Matej Balog\textsuperscript{3}, 
  Marvin Eisenberger\textsuperscript{3}, 
  Sergey Shirobokov\textsuperscript{3}, 
  Ngân Vũ\textsuperscript{3}, \\[0.5ex] %
  Martin Dixon\textsuperscript{2}, 
  Borivoje Nikolić\textsuperscript{1}, 
  Parthasarathy Ranganathan\textsuperscript{2}, 
  Sagar Karandikar\textsuperscript{1$\dagger$}
}

\thanks{\textsuperscript{$\dagger$}Work done while the author was also affiliated with Google.}

\affiliation{%
  \vspace{1ex} %
  \institution{
    \textsuperscript{1}University of California, Berkeley \\ %
    \textsuperscript{2}Google \qquad \qquad %
    \textsuperscript{3}Google DeepMind
    \vspace{1ex}
  }
  \country{} 
}
\email{raghavgupta@berkeley.edu, {bora, sagark}@eecs.berkeley.edu}
\email{{avjain, abegonzalez, mgdixon, parthas}@google.com}
\email{{anovikov, posenhuang, matejb, meisenberger, shirobokov, nganvu}@google.com}

\begin{teaserfigure}
\end{teaserfigure}

\renewcommand{\shortauthors}{Gupta et al.}

\begin{abstract}

Agile hardware design flows are a critically needed force multiplier to meet the exploding demand for compute. 
Recently, agentic generative artificial intelligence (AI) systems have demonstrated significant advances in algorithm design, improving code efficiency, and enabling discovery across scientific domains.

Bridging these worlds, we present {\alphaarchitect}, an automated \emph{computer architecture discovery} system built on {\anonevolve}. We show {\alphaarchitect}'s ability to automatically design/implement state-of-the-art (SoTA) cache replacement policies (architecting new mechanisms/logic, not only changing parameters), broadly within the confines of an established cache replacement policy design competition.
  
  In two days and without human intervention, {\alphaarchitect} generated a policy achieving a \num{5.322}\% IPC speedup improvement over the prior SoTA on public multi-core Google Workload Traces. On the heavily-explored single-core SPEC 2006 workloads, in only 18 days {\alphaarchitect} generated a policy showing a \num{0.907}\% IPC speedup improvement over the existing SoTA (a similar "winning margin" as reported by the existing SoTA). 
  Comparing against the effort involved in developing the prior SoTA policies, {\alphaarchitect} achieved these gains 3-5$\times$ faster than humans. 
  
  Agentic flows also create an opportunity once a hardware system is deployed, which we call ``post-silicon hyperspecialization''. This means having the agent tune runtime-configurable parameters exposed in hardware policies to further align the policies with a specific workload (mix). Exploiting this, we demonstrate a \num{2.374}\% IPC speedup improvement over prior SoTA on the SPEC 2006 workloads.

Since {\alphaarchitect} is a first-of-its-kind architectural discovery system, we also outline lessons learned and broader implications for computer architecture research in the era of agentic AI.
For example, we demonstrate the phenomenon of ``simulator escapes'', where the agentic AI flow discovered and exploited a loophole in a popular microarchitectural simulator---a consequence of the fact that these research tools were designed for a (now past) world where they were exclusively operated by humans acting in good-faith.

\end{abstract}

\maketitle

\section{Introduction}

\begin{figure}
  \centering
  \includegraphics[width=\linewidth]{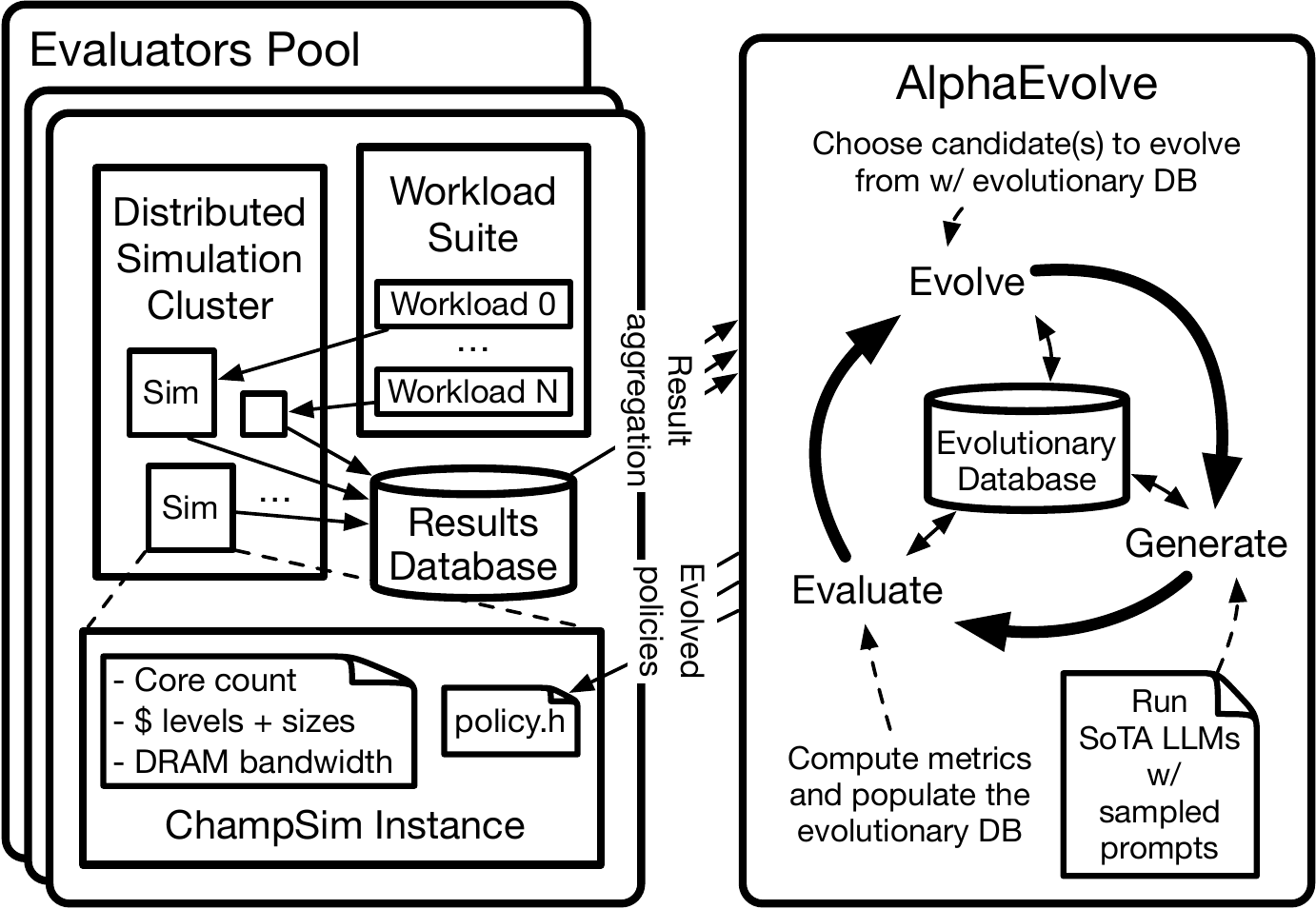}
  \caption{High-level system diagram of {\alphaarchitect}, our agentic-AI-based computer architecture discovery system. In this example, novel cache replacement policy candidates are automatically designed/implemented by {\anonevolve} in ChampSim, a popular trace-based microarchitectural simulator. ChampSim is then compiled and run with a specified workload suite (e.g., SPEC) to evaluate the new policy on a target metric (e.g., IPC). This process continues iteratively, with {\alphaarchitect} continually proposing and evaluating new logic/mechanisms within the policy.}
  \label{fig:ae-loop}
\end{figure}

The computing industry continues to battle the rift between the exploding demand for compute and the need for human-time-intensive per-domain specialization to make up for the plateau in classical hardware scaling techniques. Continued efforts in agile hardware design, more recently driven by the infusion of machine learning (ML), are welcomed to help bridge this gap. While inroads have been made in applying ML to RTL-to-chip flows, there is less focus on enabling automated ``discovery'' in earlier stages of the hardware design process, beyond using ML to explore parameterized design spaces. In this paper, we answer the following: Can modern generative AI tools help computer architects in early discovery, ideation, and pathfinding, so that we can \emph{more quickly} architect \emph{a greater number} of specialized architectures?

Recently, agentic flows based on large language models (LLMs) have emerged as useful tools to automate algorithm design, software systems design, software efficiency, and scientific discovery~\cite{cheng2025barbarians, novikov2025alphaevolve}.
Guided by high-level prompting from a human, such a flow can generate, implement, and evaluate ideas much faster than a human designer.
This shifts the designer's focus to problem formulation, creative ideation, and experimental design.

The basic premise of {\alphaarchitect} is straightforward: Agentic, evolutionary ML-based tools can automatically and iteratively express and evaluate new concepts as code, so we should have these tools write code to express new architectures in the context of our community standard (micro)architectural simulators. Closing the loop, since our simulators can provide quantitative feedback, these agents can work iteratively to improve their solutions with new mechanisms/logic developed on-the-fly by an LLM. In the rest of this paper, we discuss the benefits and tradeoffs of this approach and express a call-to-action for our community infrastructure to enable us to get the most out of these agentic AI tools.

More concretely, we make the following key contributions:

\subsubsection{The {\alphaarchitect} system} We design and implement {\alphaarchitect}, an agentic AI system that automates computer architecture discovery. {\alphaarchitect} expresses architectures as C++ models in ChampSim~\cite{gober2022championship} and builds on {\anonevolve}~\cite{novikov2025alphaevolve}, including a highly distributed evaluation setup to discover novel architectures. An outline of {\alphaarchitect} is shown in Figure~\ref{fig:ae-loop}.

\subsubsection{Automatically designing state-of-the-art cache replacement policies} We show that {\alphaarchitect} can discover state-of-the-art (SoTA) cache replacement policies, broadly adhering to the experimental design of established community cache replacement policy design championships, such as the Cache Replacement Championship~\cite{2ndcrc}, most recently held in early 2017. The discovered policy improves IPC speedup normalized to LRU over the prior SoTA policy (published in followup work~\cite{shah2022effective} in 2022) by \num{0.737}\% on single-core SPEC 2006~\cite{10.1145/1186736.1186737}  in the absence of prefetching, and by \num{0.907}\% in the presence of prefetching. These gains are similar to prior ``winning margins'' reported by championship winners/new SoTA work in this domain. Notably, comparing against the effort involved in developing the prior SoTA policies, {\alphaarchitect} achieved these gains 3-5$\times$ faster than humans. 

\subsubsection{Policy deep-dive} To demonstrate the kinds of mechanisms/logic that {\alphaarchitect} can design, we perform an ablation study on {\policyone}, one of the winning replacement policies generated by {\alphaarchitect}. We break down the novel policy mechanisms created by {\alphaarchitect} qualitatively and quantitatively. %

\subsubsection{Designing beyond SPEC} In addition to discovering new policies for SPEC workloads, which is known to not be representative of cloud workloads~\cite{10.1145/2150976.2150982, 10.1145/3695053.3731411, gan2019open}, we use the publicly-available Google Workload Traces~\cite{Google_Workload_Traces_Version_2} to discover high-performance policies for hyperscale workloads, and we show that {\alphaarchitect} can design a policy that produces a \num{5.322}\% IPC speedup improvement over prior SoTA policies within two days. Previously designing a handcrafted hyperscale-centric policy would require a concerted, human-intensive effort spanning months.

\subsubsection{Post-silicon ``hyperspecialization''} Design flows like {\alphaarchitect} represent a new paradigm in automated discovery. Although we have only explored pre-silicon discovery thus far, we can now ask the question: Could {\alphaarchitect} further improve its generated policy post-silicon, by configuring the policy specifically for each workload at runtime? While this process would traditionally be extremely human-intensive, we show that {\alphaarchitect} can further improve IPC speedup by \num{2.374}\% on average (up to \num{8.1}\% on \texttt{mcf\_46B}) over the prior state-of-the-art by customizing post-silicon runtime-configurable policy parameters on memory-intensive SPEC 2006 workloads. \emph{It is important to note that this is the only part of this work where we only tune parameters; all other uses of {\alphaarchitect} are for devising new policy logic and mechanisms.}

\subsubsection{Strengths, pitfalls, and a call-to-action} {\alphaarchitect} provides a proof-of-concept that ML-assisted design flows can enable architectural discovery. However, this leaves many open questions still to be answered. We take an experiential \emph{and} evidence-based deep-dive on the strengths and weaknesses of such systems today, including critical issues around about overfitting versus specialization, extrapolation from representative workloads, and the implications for our community evaluation infrastructure in the era of agentic AI.

\section{Background}
\label{sec:background}

To set the context, we briefly cover very recent advances in two core areas: (1) agentic LLM-based evolutionary discovery tools and (2) cache replacement policy design and competition-based evaluation. A more substantial discussion of related work can be found in Section~\ref{sec:related-work}.

\subsection{LLM-based Evolutionary Agents}
\label{sec:agents}

Recently, there has been an emergence of coding agents designed for algorithmic and scientific discovery, such as Google DeepMind's AlphaEvolve. These agents use LLMs in combination with evolutionary search to discover new solutions across varied domains such as their use in mathematical proofs~\cite{georgiev2025mathematical} and software systems research~\cite{cheng2025barbarians}.
Such agentic systems allow for the evolution of an entire code file (e.g., hundreds of lines of code) in any programming language through automatic generation and evaluation of code.
This is done by a human programmer first providing setup instructions (e.g., task directives, evaluation criteria, and background knowledge) to the system and an initial solution to seed an evolutionary database. Next, the setup instructions and prior code blocks from the evolutionary database are sampled to create prompts, which are fed to LLMs to generate new code blocks.
These new code blocks are fed into an evaluator, then ranked and fed into the evolutionary database before continuing on iteratively.
Such ranking and evolution is done by way of evolutionary algorithms such as island-based evolution~\cite{romera2024mathematical, tanese1989distributed} or MAP-Elites evolution~\cite{mouret2015illuminating}, which balance genetic diversity based on behaviors (e.g., code size and complexity) as well as performance on a specified metric (i.e., fitness score).

\subsection{Cache Replacement Championships}
\label{sec:priorwinners}

With the goal of incentivizing the community to develop novel cache replacement policies, The 1st Journal of Instruction-Level Parallelism (JILP) Cache Replacement Championship (CRC-1)~\cite{Alameldeen_Jaleel_Qureshi_2010} was held in 2010 to compare different last level cache (LLC) cache replacement algorithms in a common evaluation framework.
Given a fixed storage budget and predefined single- and multi-core configurations, competitors could submit cache replacement policies that would then be ranked by a common set of workloads.
The latest championship, CRC-2~\cite{2ndcrc}, had four tracks: (1) single-core without prefetching, (2) single core with prefetching, (3) multi-core without prefetching, and (4) multi-core with prefetching.
Each track was ranked separately, and the winner was decided based on the aggregate score across all tracks.
The single-core track score was determined by the geometric-mean of replacement algorithm speedups across a set of single-threaded workloads, while in the multi-core track, scores were determined by the weighted speedup across a set of multi-program and multi-threaded workloads. 

Looking across CRC-1, CRC-2, and other non-championship publications in the cache replacement policy domain~\cite{gao10, Wu11, mowry92, SHiP++, jain16, shah2022effective}, margins of instructions-per-cycle (IPC) improvement normalized to LRU compared to prior winners are typically reported to be in the range of 1\% to 3\%, for single-core ChampSim configurations across various mixes of workloads (including SPEC, GAP~\cite{beamer2015gap}, CVP1~\cite{Perais_Sheikh_Rotenberg_Wilkerson_Srinivasan_Alameldeen_Lipasti_2018}, and more).
For direct comparison, Mockingjay~\cite{shah2022effective}, the prior state-of-the-art cache replacement policy, saw a \num{1.6}\% and \num{1.2}\% increase in single-core performance normalized to LRU over Hawkeye~\cite{jain16} across a memory-intensive subset of SPEC 2006 workloads, without prefetching and with prefetching, respectively.
We make two observations across the winning policies:

\begin{itemize}
\item The winning margins are typically small and are getting smaller, which attests to the difficulty of the problem domain and the expectation that the headroom is shrinking.
\item Nevertheless, the field of cache replacement has progressed gradually with wins in the 1\%-3\% range accumulating over time.
However, each advance requires a concerted effort, with (usually) multiple researchers studying the problem over a long period of time. 
\end{itemize}

\section{{\alphaarchitect} System Design}
\label{sec:alphaarchitect}

{\alphaarchitect} harnesses an agentic generative AI system to automate computer
architecture discovery. At its core, we utilize the fact that a
common mechanism for expressing new computer architectures
is writing code in a software (micro)architectural simulator,
which is a good match for an evolutionary large language model-based code-authoring agent such as {\anonevolve}. {\alphaarchitect} integrates ChampSim, a widely used trace-based microarchitectural simulator written in C++, with {\anonevolve} and a distributed evaluation backend. {\anonevolve} is capable of rewriting any C++ code file in ChampSim, but for the results presented in this paper, we restrict it to designing last-level cache replacement policies. {\alphaarchitect} works iteratively: automatically implementing candidate policies in ChampSim and running large-scale, distributed ChampSim simulations to evaluate quality (primarily on instructions per cycle (IPC) achieved). The candidate policies and evaluation feedback guide an evolutionary search algorithm to propose changes for a new generation of candidates.
Figure~\ref{fig:ae-loop} shows a high-level overview of {\alphaarchitect}. In the rest of this section, we show the key components of {\alphaarchitect} in the context of using it to design new cache replacement policies.

\subsection{Prompting an LLM to be an Architecture Research Agent}

\begin{figure}
  \centering
  \begin{lstlisting}[language={}, frame=single, breaklines=true, breakindent=0pt, basicstyle=\footnotesize\ttfamily]
Act as an expert software developer and computer architect. The codebase that you are working on is a simulator for an out-of-order superscalar processor running a program trace. Your task is to iteratively improve the indicated section of the codebase, which models a replacement policy for the caches in the simulated processor and win the Cache Replacement Championship. That is, you want to design the best possible cache replacement policy for an out-of-order superscalar processor and implement it in the simulator. The primary goal is to increase the scores on the provided evaluation metrics, where larger values are better. One of these metrics is the number of instructions-per-cycle (IPC) that the processor executes. A better processor will achieve a higher IPC.

[...]

Ensure the code you introduce is realizable in hardware. Ensure you don't use more than 48KB state for your replacement policy. This is separate from the size of the cache itself. This is a strict limit and you must adhere to it honestly. Otherwise you will be disqualified.

[... context about simulator caveats/APIs, max lines to change, workloads, system configs, prior working programs, and prior literature ...]
\end{lstlisting}
  \caption{Simplified example of a prompt given to the {\anonevolve} used in {\alphaarchitect} including persona, background information, guidance.}
  \label{fig:example-prompt}
\end{figure}

To automatically generate new policies, we provide {\anonevolve} with an instruction prompt to describe the task and provide relevant background and context.
Figure~\ref{fig:example-prompt} lists a simplified example of a prompt used, that we aimed to be an expert computer architect persona.
Combinations of prior policies (such as DRRIP, SHiP, Mockingjay), literature references, expectations of IPC headroom, and simulator interface information were provided to help an ensemble of fast and efficient (Gemini 2.5-Flash) and deep reasoning (Gemini 2.5-Pro) LLMs to generate varied policies.
After each run, the prompt is automatically configured to provide solutions and metadata from previous runs to create the next evolution to test. 
Importantly, optimizations added included prompting the models to have more variation, respect hardware budget constraints (e.g., state size) required by the prior cache replacement championships, and reduce code complexity (e.g., upper bound of 1K lines of code).
This included variations of word-play to ``entice" the models to generate sufficiently different responses.
The prompt described above is templatized, with various components probabilistically sampled and inserted to generate a variety of prompts at runtime to improve the volume of ideas explored.

\subsection{Starter Code}

We provide ChampSim's default replacement policy C++ implementation files as starter code for {\anonevolve} to add new policy logic and change/remove ineffective logic.
While we experimented with simple policies as starting points (e.g., LRU), we eventually converged on providing the Mockingjay policy source code, for faster progress and convergence due to slow simulation speeds.
Similar to detailed documentation in the prompt, the file was also annotated extensively with additional API information, assumptions, and more.

\subsection{Workloads and ChampSim Hardware Configurations}

The choice of workload that can be fed to ChampSim in {\alphaarchitect} is flexible. In this case, we selected both SPEC 2006 and Google Workload Traces Version 2, representing a mix of publicly-available workloads with a variety of memory-traffic characteristics.
These SPEC 2006 traces, obtained from CRC-2, use SimPoints~\cite{perelman2003using} on multiple high LLC misses-per-kilo-instruction (MPKI) workloads~\cite{shah2022effective}.
We use the publicly-available Google Workload Traces Version 2 and convert into a ChampSim-compatible format.
For ChampSim microarchitectural configurations, we also used the default CRC-2 single- and multi-core configurations as seen in Table~\ref{tab:mem_config}.
In Section~\ref{sec:challsols}, we describe the process of running ChampSim on our distributed backend to maximize parallelism on long-running evaluations.

\subsection{Evaluation Metrics}
To match CRC-2, {\alphaarchitect} optimizes for the IPC of the cores running the given workloads.
For single-core ChampSim configurations, this amounts to the geometric-mean IPC speedup over the baseline ChampSim LRU policy.
For multi-core configurations, this is the weighted IPC speedup over the same LRU policy.

While the initial LLM prompt was given a hardware budget and tips to improve explainability of generated responses, the underlying LLMs occasionally ignored instructions given.
This resulted in combinations of unrealistic hardware implementations, circumventing the championship constraints, and complex and difficult to understand policies.

In addition to stronger prompt verbiage (e.g. explicit lines of code wanted), we added the ability to measure the number of lines of code generated and negatively rewarded the system for longer responses.
Once a final candidate policy was identified (i.e., after sufficient iterative progress on target metrics), additional manual checking was also done to verify that output policies represented realistic, hardware-implementable designs.

{
\sisetup{round-mode=places,round-precision=1}
\begin{table}
  \centering
  \caption{Simulated ChampSim memory system configurations obtained from the 2nd Cache Replacement Championship.}
  \label{tab:mem_config}
  \begin{adjustbox}{width=\columnwidth,center}
  \begin{tabular}{llcccc}
    \toprule
    \multicolumn{2}{c}{\multirow{3}{*}{\textbf{Parameter}}} & \multicolumn{2}{c}{\textbf{Single-Core Config.}} & \multicolumn{2}{c}{\textbf{Multi-Core Config. (4 Cores)}} \\
    \cmidrule(lr){3-4} \cmidrule(lr){5-6}
     & & \multicolumn{2}{c}{\textbf{Prefetch}} & \multicolumn{2}{c}{\textbf{Prefetch}} \\
     \cmidrule(lr){3-4} \cmidrule(lr){5-6}
     & & \textbf{Disabled} & \textbf{Enabled} & \textbf{Disabled} & \textbf{Enabled} \\
    \midrule
    Cache Sizes & L1 Data & 48 KiB & 48 KiB & 48 KiB & 48 KiB \\
    & L1 Instruction & 32 KiB & 32 KiB & 32 KiB & 32 KiB \\
    & L2 (Per Core) & 512 KiB & 512 KiB & 512 KiB & 512 KiB \\
    & LLC (Shared) & 2 MiB & 2 MiB & 8 MiB & 8 MiB \\
    \midrule
    Prefetchers & L1 Data & - & Next Line & - & Next Line \\
    & L2 (Per Core) & - & PC Stride & - & PC Stride \\
    \midrule
    DRAM & Bandwidth & \num{25.6} GB/s & \num{25.6} GB/s & \num{25.6} GB/s & \num{25.6} GB/s \\
    \bottomrule
  \end{tabular}
  \end{adjustbox}
\end{table}
}

\subsection{Challenges in Using a Microarchitectural Simulator as an AI Evaluator}
\label{sec:challsols}

Evaluation of each policy change takes a significant amount of time because microarchitectural simulations are slow and ChampSim itself is single threaded.
For example, the evaluation for single-core SPEC 2006 using ChampSim in the CRC-2 competition framing requires running workload traces for 1B instructions.
Depending on the host machine running a simulation, the ChampSim configuration, and the benchmark being simulated, this can result in a single simulation taking over 12 hours on a SoTA server-class CPU.
Multi-core evaluations are even slower, often taking 2-4 days for some workload mixes since simulation is single threaded, resulting in at least a proportional increase in runtime relative to the number of cores simulated.

To reduce {\alphaarchitect}'s overall iteration time and avoid overfitting, we reduced the number of instructions run when evaluating proposed policies in each iteration of the evolutionary discovery process.
While this resulted in some generalization issues---where results on shorter runs did not generalize to longer, representative runs---we took a cascaded approach wherein policies were created on shorter runs (i.e., 100M instructions) then later validated on longer runs (e.g., 1B instructions for SPEC results to match competition framing).

To maximize iteration speed, we also added the capability to run parallel long-running simulations on a distributed cluster. As we began to run microarchitectural simulations in this fashion, we experienced issues common to building distributed systems, including seeing simulations get canceled due to maintenance operations (e.g., kernel updates) and other hardware instability.
Typical scale-out software workloads would avoid this issue by periodically checkpointing intermediate state to persistent storage to restart from.
However, the ChampSim simulator does not support this functionality, resulting in failed multi-day simulations (used for validating proposed policies) having to be restarted from scratch. 
Thus, additional infrastructure was built to automatically restart jobs and a locally managed persistent cluster was set up to handle long running jobs susceptible to interruption.
In Section~\ref{sec:discussion-eval} we discuss augmenting computer architecture evaluation infrastructure to avoid these issues.

\subsection{Putting Together the Pieces}

After the above inputs are provided and the environment configured, {\alphaarchitect} controls the discovery process, automatically generating novel policies, testing them against the evaluation environment, and continually improving them using evolutionary search guided by quantitative evaluator feedback.
Due to prompt randomization, LLM entropy, and evolutionary algorithm sampling, evolution speed varies as the system explores the design space and reaches a state-of-the-art solution.

\begin{table*}[t] %
  \centering
  \caption{Comparison of Policy Evolution Setups. In \texttt{Evaluation Metric}, \texttt{geomean} is geometric-mean across benchmarks while \texttt{mean(IPC)} refers to the arithmetic mean of IPC across all cores.}
  
  \label{tab:policy_evolution_wide_simplified}

  \begin{tabular}{cccccc}
    \toprule
    \multirow{2}{*}{\textbf{Policy}} &
    \multirow{2}{*}{\textbf{Workload}} &
    \multicolumn{2}{c}{\textbf{Simulation Instructions}} &
    \multirow{2}{*}{\textbf{Evaluation Metric}} &
    \multirow{2}{*}{\textbf{Evolution Time}} \\
    \cmidrule(lr){3-4}
     &  & \textbf{Evolution} & \textbf{Validation} &  &  \\
    \midrule
    {\policyone} & \makecell[c]{19 Memory Intensive\\SPEC06 Benchmarks} & \makecell[c]{max(50M, 100M)\\within 1 hr} & 1B &  $geomean\left(\frac{\text{IPC}_{\text{policy}}}{\text{IPC}_{\text{lru}}}\right)$ & 18 days \\
    \addlinespace
    {\policyonept} & \makecell[c]{19 Memory Intensive\\SPEC06 Benchmarks} & 1B & 1B & $geomean\left(\frac{\text{IPC}_{\text{policy}}}{\text{IPC}_{\text{lru}}}\right)$ & \textless{} 8 days \\
    \addlinespace
    {\policythree} & 11 Google Workload Traces & 50M & 75M & $geomean\left(\frac{\text{mean(IPC}_{\text{policy}})}{\text{mean(IPC}_{\text{lru}})}\right)$ & 4 days \\
    \addlinespace
    {\policyfour} & 11 Google Workload Traces & 20M & 75M & $geomean\left(\frac{\text{mean(IPC}_{\text{policy}})}{\text{mean(IPC}_{\text{lru}})}\right)$ & 2 days \\
    \bottomrule
  \end{tabular}
\end{table*}

\section{Using {\alphaarchitect} to Automatically Design LLC Replacement Policies for Single-Core Systems}

In this case study, we use {\alphaarchitect} to design a last-level cache replacement policy, {\policyone}, to compete in the single-core portion of the CRC-2.

\subsection{Methodology}
\label{sec:sc-methodology}

{\policyone} is constructed by {\alphaarchitect} optimizing for SimPoint~\cite{perelman2003using} traces from the SPEC 2006 workload suite on single-core ChampSim system configurations with both prefetching and no prefetching, as shown in Table~\ref{tab:mem_config}.  To align with recent cache replacement policy evaluations~\cite{shah2022effective}, we only use memory-intensive SPEC workloads that have LLC MPKI \textgreater{} 1 with the LRU replacement policy. Workload-level speedup is measured as the IPC ratio of the generated policy over an LRU baseline ($\frac{IPC_{policy}}{IPC_{lru}}$) while suite-level speedup is measured as the geometric-mean of workload-level speedups as seen in Table~\ref{tab:policy_evolution_wide_simplified}.

For speedy evolution and to prevent overfitting, we provide feedback to {\alphaarchitect} by simulating a relatively small number of instructions (i.e., upto 100M instructions for each individual workload trace) to allow the simulations to complete within one hour. However, for validation, we collect final results by executing 1B instructions for each workload trace to comply with the CRC-2 competition framing.
Quicker feedback allows {\alphaarchitect} to explore more solutions in the given time frame (further discussed in Section~\ref{sec:discussion-eval}).

\subsection{{\policyone} Description} \label{sec:policyone-desc}

\begin{figure}
  \centering
  \begin{lstlisting}[language=codediff, frame=single, breaklines=true, breakindent=0pt, basicstyle=\footnotesize\ttfamily]
+ // --- HAWKS AND DOVES STATE ---
+ // A bit-packed 2b sat. ctr per cache line
+ // Tracks usage intensity and follows state budget
+ // Each byte holds 4 counters
+ std::vector<uint8_t> packed_usage_counter;

/* find a cache block to evict */
long policy31::find_victim(...) {
  for (...) {
+    // Retrieve H&D usage and use for ETR
+    current.usage = get_usage(set, way);
+    current.effective_etr = abs(current.etr_val) - (current.usage * BONUS_PER_USE);
  }
}

/* called on every cache hit and cache refill */
void policy31::update_replacement_state(...) {
+  if (hit) {
+    // Inc. usage counter (saturating at 3)
+    // This makes frequently-used blocks stickier
+    increment_usage(set, way);
+  } else {
+    // Fill on a miss (a new line inserted)
+    // The new block starts with a usage of 0
+    reset_usage(set, way);
+  }
}
\end{lstlisting}
  \caption{Example {\policyone} modifications in the form of a diff to implement the Hawks and Doves mechanism. The \texttt{packed\_usage\_counter} and corresponding \texttt{get}-, \texttt{increment}-, and \texttt{reset\_usage} setter/getters are used to help determine eviction candidates.}
  \label{fig:handd}
\end{figure}

Starting from the Mockingjay codebase, {\alphaarchitect} experimented for 18 days by adding (or changing/removing) code for new policy logic/mechanisms to create {\policyone}.

Mockingjay predicts the future reuse distance of each line using a PC-based predictor and then evicts the line whose predicted reuse is furthest in the future.  {\alphaarchitect} augmented the policy with several new techniques discussed below.

\subsubsection{Insertion Quality}

{\policyone} includes an Insertion Quality Predictor (IQP) to identify PCs that bring in dead blocks (i.e., blocks that are never used), and it penalizes reuse distance prediction for dead blocks by inflating their predicted reuse distance (which makes them more likely to get evicted). 

\subsubsection{Hawks and Doves} 

{\policyone} tracks the usage intensity of each block with a 2-bit saturating counter. Blocks with a higher usage count are considered more valuable and are less likely to be evicted. Figure~\ref{fig:handd} shows the logic to adjust ETR (Estimated Time of Reuse) based on usage.

\subsubsection{Prefetch-Aware Retention}

{\policyone} introduces logic to deprioritize easy-to-prefetch sources by giving them a high ETR and prioritize difficult-to-prefetch sources by giving them a low ETR. 

\subsubsection{Cache Pressure-Aware Adaptive Throttling (CPAAT)} 

{\policyone} introduces a new mechanism to dynamically adjust the bypass aggressiveness based on the overall miss rate (i.e., cache pressure).
Under high pressure, it's more conservative with new insertions, preserving the existing working set.

\subsection{Results}
\label{sec:spec-baseline}

\begin{figure*}[t]
  \centering
  \includegraphics[width=\textwidth]{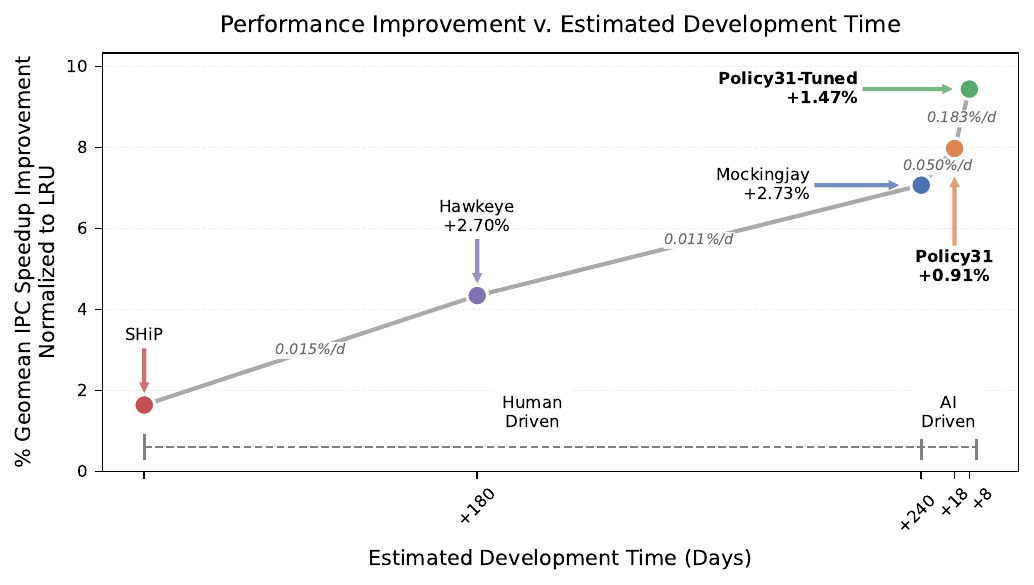}
  \caption{Performance improvement (suite-level geomean IPC speedup normalized to LRU) compared to estimated development time of replacement policies for the single-core prefetch-enabled ChampSim configuration on memory-intensive SPEC06 workloads. Slope (grey, italics) denotes percentage point improvement per day.}
  \label{fig:improvement_by_time}
\end{figure*}

\begin{figure*}
  \centering
  \includegraphics[width=\textwidth]{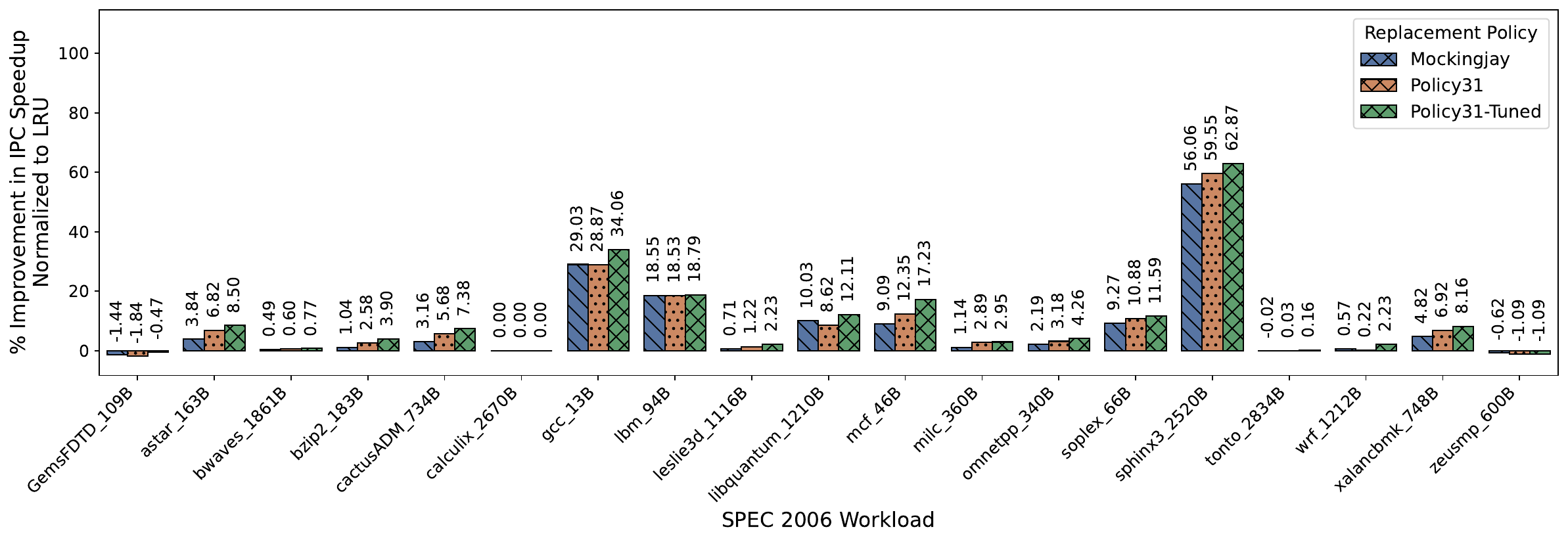}
  \caption{Per-workload improvement in policy IPC speedup normalized to LRU for the single-core prefetch-enabled ChampSim configuration running SPEC 2006 memory intensive workloads.}
  \label{fig:policyone-param-tune-1c2a1b-all-bmarks}
\end{figure*}

We compare suite-level IPC speedup improvement normalized to the standard LRU baseline. We find that {\policyone} improves IPC speedup normalized to LRU by \num{12.2}\% and \num{8.0}\% in the no prefetch and prefetch configurations, respectively (versus \num{11.4}\% and \num{7.0}\% for Mockingjay).
Figure~\ref{fig:policyone-param-tune-1c2a1b-all-bmarks} shows the workload-level improvement across the suite in the prefetch-enabled case, demonstrating overall improvement and no significant outliers.

To put these improvements in context, Section~\ref{sec:priorwinners} describes that the historical rate of improvement for cache replacement for single-core ChampSim configurations is reported to be in the 1-3\% range per solution.
Each improvement in prior work has needed significant effort by multiple researchers over several months, whereas here we have demonstrated that {\alphaarchitect} can achieve similar gains in less than 3 weeks.

Figure~\ref{fig:improvement_by_time} compares the performance improvement achieved by successive SoTA cache replacement policies compared to the estimated time of developing these policies for the single-core prefetch-enabled ChampSim configuration on memory-intensive SPEC06 workloads. The estimated development time was obtained from their creators. These reported times include only time to develop the policies, excluding, for example, paper writing time. When comparing the effort involved in developing {\policyone} and the prior SoTA policies, we find that {\alphaarchitect} achieved its gains 3-5$\times$ faster than humans.
Furthermore, replacement policy development for SPEC workloads is well explored as prior work has demonstrated that existing solutions come within 90\% of the hit rate optimal solution~\cite{shah2022effective}. This points to the increasing difficulty of finding performance improvements for this problem domain.

\subsection{Ablation Study}

\begin{figure}
  \centering
  \includegraphics[width=\linewidth]{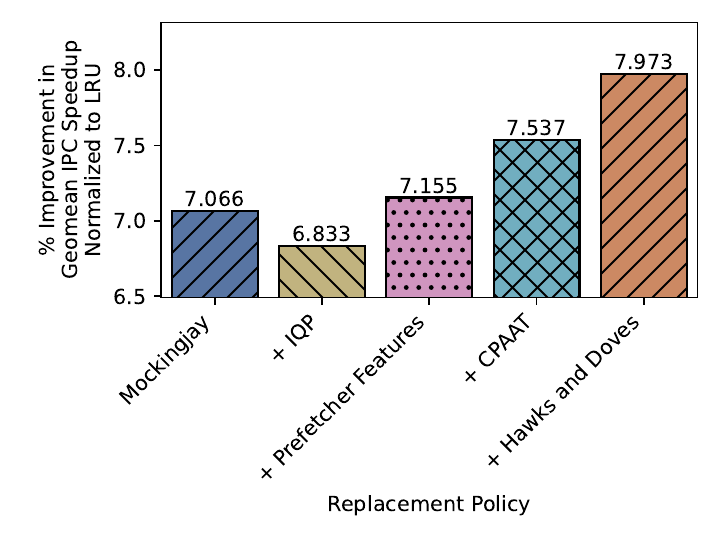}
  \caption{Ablation study measuring improvement in suite-level geomean IPC speedup with each new technique that composes {\policyone} for the single-core prefetch-enabled ChampSim configuration running SPEC06 memory intensive workloads.}
  \label{fig:policyone-1c2a1b-ablation}
\end{figure}

To better understand the source of performance improvement for {\policyone}, we run an ablation study breaking down the impact of key policy features created by ArchAgent (described in ~\ref{sec:policyone-desc}).
Figure~\ref{fig:policyone-1c2a1b-ablation} shows the ablation study results when running on the default single-core prefetch-enabled ChampSim configuration running workloads from SPEC 2006 suite for 1B instructions.
Selectively, the study re-enables the various mechanisms that comprise {\policyone} one-by-one, until the complete {\policyone} is evaluated.
As seen in the figure, most of the improvement is coming from CPAAT and Hawks and Doves, providing a \num{0.818}\% improvement, combined.
IQP and prefetch management result in marginal gains and only show benefits when combined with all other mechanisms.

This highlights a challenge with machine-generated policies, where manual work is still needed to not only validate the performance improvement but also understand the root cause.
To have high confidence in machine-generated policies, the process of ablation and verification will remain a key bottleneck that could also benefit from automation, as discussed further in ~\ref{sec:discussion-llm-verif}. 

\section{Using {\alphaarchitect} for Runtime-Configurable, Workload-Specific Hyperoptimization}
\label{sec:workload-spec-opts}

We observe that {\alphaarchitect}-generated policies such as {\policyone} introduce new microarchitectural techniques that make the policies flexible and amenable to runtime tuning.
These runtime parameters are expressions in the policy that do not affect storage size and are instead either constants or simple arithmetic expressions derived from constants.
We identify 13 runtime parameters in {\policyone} that fall into the following two categories: 
\begin{itemize}
    \item Constants used in scores such as ETR or bonuses/penalties for particular categories of accesses (e.g., prefetches).
    \item Thresholds such as those used to detect long ETR blocks or initiate adaptive decay of cache counters.
\end{itemize}

\subsection{Methodology}

Since we are exploring runtime tuning, both evolution and validation are performed on the memory intensive SPEC 2006 workload subset mentioned in Section~\ref{sec:sc-methodology} running for 1B instructions (Table~\ref{tab:policy_evolution_wide_simplified}). {\alphaarchitect} optimizes each workload individually and we enforce that it only changes runtime parameters instead of the entire replacement policy.

\subsection{Results}

{\alphaarchitect} creates a new runtime parameter-tuned variant of\\{\policyone}, called {\policyonept}, tuned for individual SPEC 2006 workloads on the championship single-core ChampSim prefetch-enabled configuration.
When comparing suite-level speedups,\\{\policyonept} achieves an additional \num{1.467}\% overall geometric-mean IPC improvement as compared to {\policyone} and \num{2.336}\% overall improvement compared to Mockingjay, normalized to LRU.
When analyzing workload-level speedups, Figure~\ref{fig:policyone-param-tune-1c2a1b-all-bmarks} shows that workloads such as \texttt{gcc\_13B} and \texttt{mcf\_46B} show \textgreater \num{5}\% gain while others show limited improvement (e.g., \texttt{calculix\_2670B}).

To put these results in context, Figure~\ref{fig:improvement_by_time} shows that {\alphaarchitect}'s rate of improvement with {\policyonept} is over 10$\times$ faster than prior SoTA policies. Notably, this improvement was achieved in less than eight days.

These results highlight the potential of AI-driven runtime specialization of hardware to specific workloads and we envision that such post-silicon hyperoptimizations could be commonplace, similar to automatic profile-guided optimizations prevalent in hyperscalers~\cite{chen2016autofdo, litz2022crisp, 10.1145/3373376.3378498}.

\section{Using {\alphaarchitect} to Automatically Design Multi-Core LLC Replacement Policies for Google Workload Traces}

With {\policyone}, we show {\alphaarchitect} can win within an established competition environment on SPEC 2006. 
However, it is widely established that SPEC 2006 is not representative of hyperscale cloud workloads~\cite{10.1145/2150976.2150982}. For example, SPEC workloads have much smaller instruction footprints and significantly lower instruction cache pressure as compared to hyperscale workloads. While Mockingjay shows clear wins on both single- and multi-core SPEC, Figure~{\ref{fig:policythree-4c8a1b}} shows it performs much worse than even LRU on Google Workload Traces.

Thus, in this case study, we use {\alphaarchitect} to generate multi-core cache replacement policies, {\policythree} and {\policyfour}, specialized for publicly-available Google Workload Traces Version 2.

\subsection{Methodology}

{\policythree} and {\policyfour} are generated by running 11 workloads from the Google Workload Traces Version 2 suite across both prefetch and non-prefetch multi-core ChampSim configurations stated in Table~\ref{tab:mem_config}. 
The DynamoRIO trace scheduler~\cite{1191551}, similar to an operating system scheduler, is used to schedule thread-level traces from each workload onto the four cores of the multi-core system.

Final validation is done by simulating the two policies for 75M instructions as seen in Table~\ref{tab:policy_evolution_wide_simplified}.
Here, workload mix speedup is computed as the arithmetic mean of IPC across cores normalized to the arithmetic mean of IPC across cores using the LRU replacement policy. 
We find this to be a suitable metric because all four cores are executing threads from the same workload.
Suite-level speedup is measured as the geometric-mean of workload-level speedups.

For speed, we provide evolution feedback by simulating 50M and 20M instructions for {\policythree} and {\policyfour}, respectively. The feedback metric averages suite-level speedup scores for both multi-core configurations. 

We provide the Mockingjay source code as the
starter code for both policies, and we allow {\alphaarchitect} to change the entire
replacement policy code.

\subsection{Policy Description}

Two different {\alphaarchitect} runs on these traces resulted in two vastly different policies, but both are building on the premise that code characteristics of hyperscale workloads are different, so the use of PC as a feature needs to be revisited. 

\paragraph{{\policythree}}

{\policythree} was generated over the course of four days. It maintains the core algorithm of Mockingjay, but makes one critical change: it enriches the signature used for prediction with information about the path taken to reach the current instruction as seen below. 

  \begin{lstlisting}[language={}, frame=single, breaklines=true, breakindent=0pt, basicstyle=\footnotesize\ttfamily]
// XOR prev. hist. with PC to make a uniq. signature
core_pc_history[cpu] = ((history << 1) | (history >> 63)) ^ instr_pc;
\end{lstlisting}

This is expected to help in scenarios where the calling context of a PC is important for its caching behavior. For example, a generic \texttt{memcpy} routine might be called to copy small, frequently-reused data structures in one part of a program, and to perform large, non-temporal streaming copies in another. A predictor that only looks at the PC of the \texttt{memcpy}'s load/store instructions will conflate these distinct behaviors, leading to an average prediction that is optimal for neither case.

Similar ideas have been proposed in the literature before~\cite{shi19, mirbagher2020chirp}, but not in the context of Mockingjay.

\paragraph{{\policyfour}}

{\policyfour} was generated over the course of two days and uses a completely different approach from Mockingjay. It first removes all the key components of Mockingjay (including reuse prediction and eviction based on estimated time of reuse) and then evolves into something that is very close to SHiP, but is much more adaptive and sensitive to larger code footprints. 

In particular, two key ideas make {\policyfour} different:

\begin{itemize}
    \item {\it Tagged Predictor Table}: PC-based cache predictors usually allow for aliasing, and the tables are simply indexed by a hash without any tag matching. {\policyfour} explicitly stores a 3-bit tag inside the predictor entry. Predictions are retrieved only if the tags match. If the tags don't match, it resets the entry. This prevents ``destructive aliasing," making the predictor more precise.
    \item {\it Learning Signal}: SHiP typically trains its predictor at the time of eviction. When a block is kicked out of the cache, SHiP checks whether the block was used. If yes, the counter for the PC that inserted it is incremented. If not, the counter is decremented. By contrast, {\policyfour} updates the predictor when the line is accessed. If the line hits, the counter for the PC corresponding to the access is incremented. If the line misses, the counter for the PC corresponding to the access is decremented.
\end{itemize}

While subtle, the second difference is significant because it allows {\policyfour} to learn much quicker than SHiP or Mockingjay. Waiting for cache eviction to learn can prolong learning feedback.  The key idea here is that some PCs have a small working set that can be cached, while most others will not see reuse due to the large working set. Thus, PCs that miss frequently (and fetch a lot of data) are penalized and PCs that hit frequently are prioritized for cache residency.

\begin{figure*}[htbp]
  \begin{minipage}{0.48\textwidth}
    \centering
    \includegraphics[width=\linewidth]{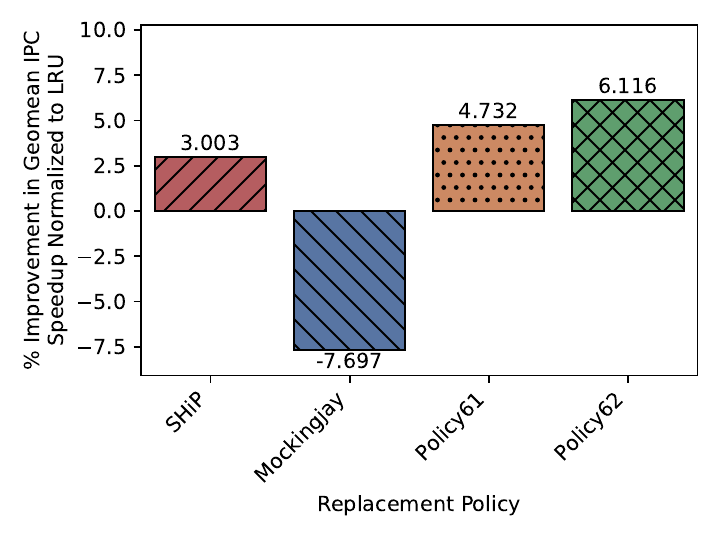}
    \caption{Improvement in geomean IPC speedup normalized to LRU for the multi-core prefetch-disabled ChampSim configuration running Google Workload Traces.}
    \label{fig:policythree-4c8a0b}
  \end{minipage}
  \hfill %
  \begin{minipage}{0.48\textwidth}
    \centering
    \includegraphics[width=\linewidth]{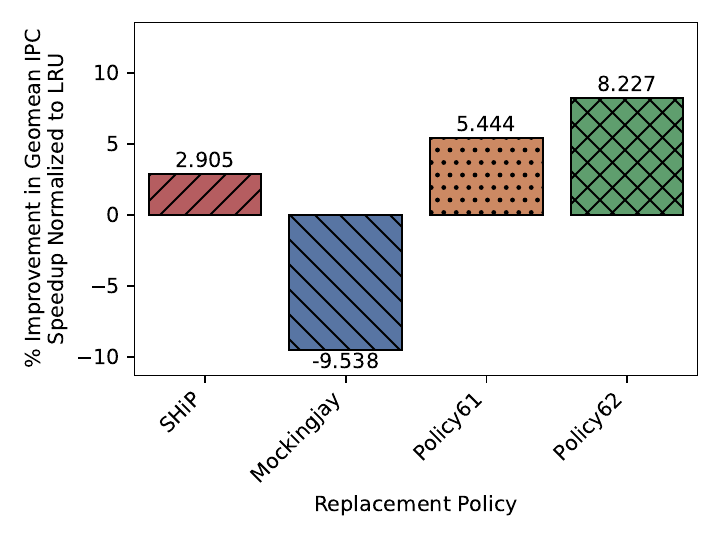}
    \caption{Improvement in geomean IPC speedup normalized to LRU for the multi-core prefetch-enabled ChampSim configuration running Google Workload Traces.}
    \label{fig:policythree-4c8a1b}
  \end{minipage}
\end{figure*}

\begin{figure*}[htbp]
  \centering
  \includegraphics[width=\textwidth]{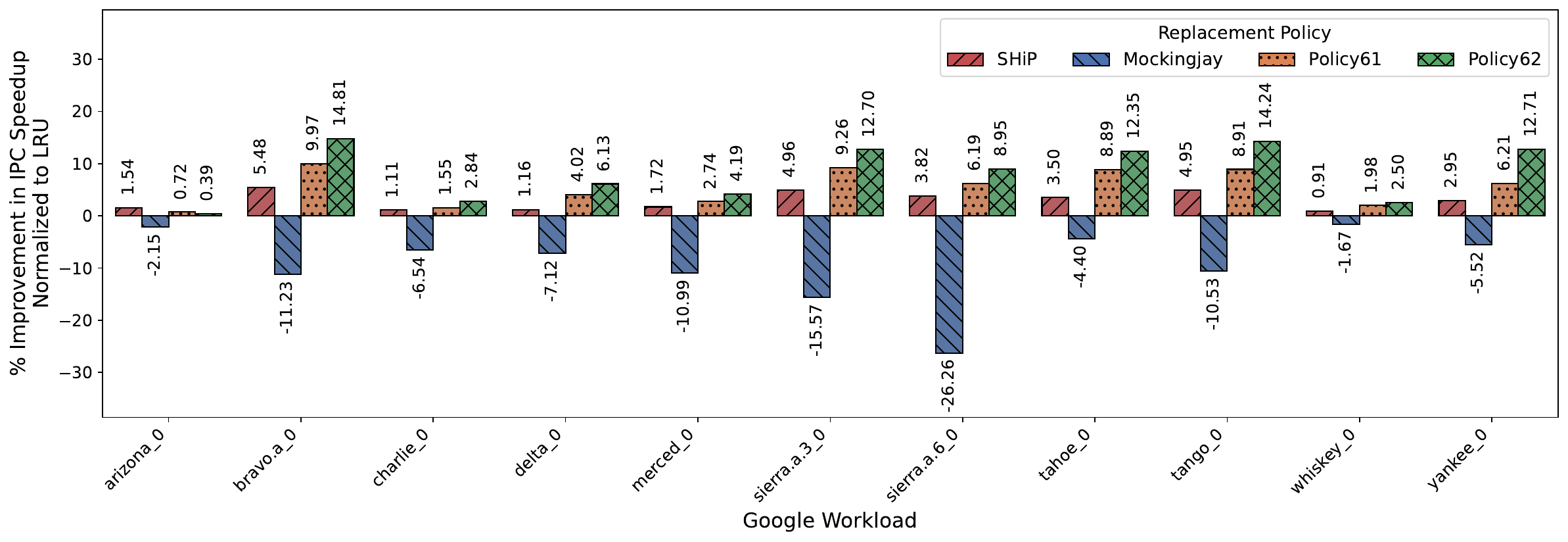}
  \caption{Per-workload improvement in policy IPC speedup normalized to LRU for the multi-core prefetch-enabled ChampSim configuration running Google Workload Traces.}
  \label{fig:policythree-4c8a1b-all-bmarks}
\end{figure*}

\subsection{Results}

Figures {\ref{fig:policythree-4c8a0b}} and {\ref{fig:policythree-4c8a1b}} compare suite-level speedups on multi-core configurations with prefetching disabled and enabled, respectively. We find a major inversion in trends here with Mockingjay performing much worse than even LRU and SHiP emerging as the prior SoTA policy. 

\begin{itemize}
    \item On the non-prefetch configuration, we find that SHiP, {\policythree}, and {\policyfour} achieve a \num{3.003}\%, \num{4.732}\%, and \num{6.116}\% improvement over LRU, respectively. 
    \item On the prefetch-enabled configuration, we find that SHiP, {\policythree}, and {\policyfour} achieve a \num{2.905}\%, \num{5.444}\%, and \num{8.227}\% improvement over LRU, respectively. 
    \item On the prefetching-enabled configuration, Mockingjay, the starter code for {\alphaarchitect} policies, shows a slowdown of \num{9.538}\% over LRU. Thus, {\alphaarchitect} recovers a performance deficit of \num{17.765}\% with a simpler policy.
\end{itemize}

Figure {\ref{fig:policythree-4c8a1b-all-bmarks}} shows workload-level speedups across the Google Workload Traces suite in the prefetch-enabled case with {\policythree} and {\policyfour} showing consistently strong performance compared to prior work. 

In our understanding, the trend inversion between Mockingjay, SHiP, and LRU, and the strong performance of {\policythree} and {\policyfour}---simpler policies derived from Mockingjay---can be attributed to the unique characteristics of these hyperscale workloads, such as deep call stacks, high degree of multithreading, and high context switch rates.

We show that {\alphaarchitect} enables designers to rapidly customize replacement policies to previously unexplored workload classes.

\section{Discussion, Future Work, and a Call-to-Action}
\label{sec:discussion}

The use of evolutionary coding agents in computer architecture research shows great promise, but will require significant improvements to our community research infrastructure, as well as continued experimentation with AI tools. In this section, we outline some of the key takeaways and potential next steps in this area.

\subsection{LLM Capabilities and Constraints}
\label{sec:discussion-llm-verif}

A central finding of this study is the remarkable capability of LLMs to generate new microarchitectural designs, including both microarchitectural technique discovery and parameter optimization.
Given a rich context, our evolutionary agent automatically generated novel, complex, and functionally correct hardware policies that improved on the performance achieved by state-of-the-art human-designed policies for both classical workloads (SPEC) and hyperscale workloads (Google Workload Traces).

Given the current approach however, confirming the realizability of generated policies remains a human intensive process, due to the lack of ASIC quality-of-result data (e.g., frequency and area) in most microarchitectural simulators.
In our case, while we found specialized prompts to be effective at generally guiding the LLMs along these lines, significant human-driven verification was still needed to guarantee compliance.
For example, to avoid complex, difficult-to-understand policies with thousands of lines of changes, we combined the use of
additional prompting to limit code size (a proxy for code complexity and hardware realizability) as well as a rule-based verifier that calculated the number of total/added lines of code and discarded samples violating constraints.
This significantly improved the explainability of solutions found, reducing the time-consuming manual ablation studies needed to understand any improvement and estimate its true hardware overhead.
However, writing rule-based verifiers is not feasible for all constraints.
As an example, inferring physical storage size from a microarchitectural model is non-trivial and thus we still rely on prompting and manual pruning/assessment for such a constraint.
This highlights a critical and interesting avenue for future work: developing principled, automated methods to inject these hard architectural constraints directly into the design process, moving beyond simple prompting or post-generation manual verification.

Additionally, despite the high quality of generated solutions evidenced in this work, there remains significant opportunity in pushing the limits of the type of solutions {\alphaarchitect} can generate. For one, there is an opportunity to collect a dataset representing a much greater sample of prior art in the cache replacement space, both in terms of implementations (software code, RTL) and descriptions (documentation, academic papers, industrial reports). Adding additional agents in the loop is another opportunity. Each of a group of multiple agents can employ a different persona, e.g., a microarchitect, an SRAM design/process design expert, a physical design expert, and a workload expert, each of which are responsible for providing different (and perhaps competing) feedback on the design, potentially using industry standard EDA tools.

\subsection{Hardware/Software Co-Design For Hyperspecialization}

The automated nature of evolutionary coding agents will have significant implications for computer architecture and systems research.
Historically, the immense human effort and simulation time required to design and validate even a single new heuristic has biased the field toward generalist solutions; policies that perform adequately, but rarely optimally, across all workloads.
The sheer productivity of an automated parallelizable agent, which can generate and evaluate thousands of policy variants in a short time, shatters this barrier and makes specialization computationally feasible, especially if deployed systems support experimentation.

This capability can usher in a new era of hardware-software co-design, moving beyond the ``one-size-fits-all" paradigm.
Instead of a static, general-purpose cache policy, a system built with even greater runtime configurability (vs. the simpler runtime parameter tuning we explored in Section~\ref{sec:workload-spec-opts}) could deploy purpose-built heuristics per workload.
For example, a heuristic evolved specifically for a critical database, a high-priority AI model, or a latency-sensitive video pipeline could yield significant efficiency benefits.
This requires the right low-cost flexible hardware-software interfaces that allow an operating system or hypervisor to securely and efficiently specialize hardware components, a promising and critical direction for future systems research.

\subsection{A Call-to-Action: Evaluation Methodology Improvements}
\label{sec:discussion-eval}

Evaluation methodology, namely computer architecture simulator performance and quality, needs to be re-evaluated in the new era of (potentially adversarial) AI agent-driven execution.

In our early experiments, we found that {\alphaarchitect} was ``willing'' to take any shortcut made available by the simulation environment, unlike a human researcher acting in good faith. We describe one such ``simulator escape''---a situation where the agentic AI flow discovered and exploited a loophole in the simulator used for evaluations. 

The ChampSim simulator does not support write-bypassing in the LLC. Verification that a policy is not performing write-bypassing only takes place \emph{in an assertion that is eliminated by the compiler when ChampSim is compiled with optimizations enabled}. Given the simulator performance concerns described in Section~\ref{sec:challsols} and below, building with these optimizations enabled is crucial.
In this case, {\alphaarchitect} developed a policy called \texttt{Policy12}, that appeared to be beating Mockingjay on the SPEC06 workloads in the single-core no-prefetch and prefetch scenarios by 3\% and 4\%, respectively. 
However, {\alphaarchitect} won by taking advantage of the fact that a bypassed write in the LLC would disappear from the system entirely. This not only avoided evicting a potentially more useful line, \emph{but also eliminated the corresponding DRAM write pressure entirely}, drastically improving IPC. Since there is no notion of ``correct computation'' in such a trace-driven simulator, there was no further way to detect this issue.

Building simulators that are either better verified or closer to the actual hardware design (e.g., RTL simulation or hardware-accelerated emulation) and thus less susceptible to ``abuse'' by the AI agents is crucial. Incorporating simulators that are closer to the actual hardware design (e.g., RTL-derived) would also have the added advantage of providing additional feedback to {\alphaarchitect}-like tools, including frequency, area, and power data collected from ASIC EDA tools.

As discussed briefly in Section~\ref{sec:challsols}, simulation speed also limits {\alphaarchitect}'s rate of discovery.
{\alphaarchitect}'s ability to reach creative solutions is inversely proportional to the evaluation latency.
This becomes untenable for long, complex simulations which are commonplace in computer architecture.
For example, evaluating a single design point for a multi-core system using ChampSim, executing hundreds of millions of instructions, can require a 12- to 24-hour turnaround.
An evolutionary search, which needs to evaluate thousands of candidates for enough variation, can be severely limited by the simulation speed, reducing the chances for the tool to try sufficiently interesting and ``risky'' ideas.
Similarly, this simulation speed also hinders evaluation of longer representative workloads (e.g., instead of simulating 1B instructions, simulating hundreds of billions of instructions representative of production programs).

As established in this discussion, to truly unleash the power of AI-driven architecture discovery, computer architecture simulators must become both more accurate and orders of magnitude faster, warranting more research into higher fidelity frameworks and simulators. 
Hardware-accelerated simulators such as  FireSim~\cite{karandikar2018firesim} could provide solutions to this problem.

\section{Related Work}
\label{sec:related-work}

There has been decades of research in designing efficient cache replacement policies and applying broader automated techniques for hardware-software discovery.

\subsection{Designing Cache Replacement Policies}

Prior work on cache replacement policy design can be viewed from the lens of different design methodologies, which can be broadly categorized into three main paradigms: handcrafted heuristics, ML-based policies, and evolutionary parameter tuning.%

\subsubsection{Heuristic-based Replacement}
Over three decades, handcrafted heuristics have changed from simple, static rules to highly sophisticated, adaptive algorithms. Early work proposed static heuristics~\cite{o93lru, smaragdakis99, wong00, lee99, karedla94, gao10, qureshi07, seshadri12, jaleel10} to avoid pathologies of LRU and LFU. More recent heuristics take a predictive approach~\cite{lai01, kaxiras01, khan10, Wu11, jain16, teran16, Jimenez17, hu02, abella05, takagi04, keramidas07,
duong12, kharbutli08, liu08, Faldu17}. They leverage past behavior to predict future caching priorities. The Mockingjay policy~\cite{shah2022effective} use predicted reuse distances to mimic Belady’s optimal caching solution~\cite{belady66}. More recently, Mostofi et al. use offline profiling to determine Insertion and Promotion Vectors for an unseen trace, however, this solution does not outperform Mockingjay~\cite{mostofi2025light}.
We use Mockingjay as the initial code that the AI agent starts from for better performance.

Another line of work recognizes that cache replacement policies must operate in the context of the system they are in.
For example, prefetch-aware policies~\cite{Wu11pacman, jain18, yuan2025joint} recognize the distinction between demand and prefetch accesses to make replacement decisions; other policies make a similar distinction for instruction accesses~\cite{mostofi2025light} or TLB accesses.
There is also work that recognizes the need for adapting replacement policies to the nature of the cache hierarchy, such as inclusive caches~\cite{jaleel2010achieving} or sliced caches~\cite{sweta2025drishti}.

From a methodological perspective, all these solutions are based on human intuition and insight.
These works do not leverage any automated methods to search the design space of replacement policies.

\subsubsection{ML-based Replacement}

One branch of research deploys ML models directly in hardware, ranging from lightweight online learning~\cite{zhou2022end} to full deep learning inference~\cite{liu20, vietri2018driving}.
For example, PARROT~\cite{liu2020imitation} casts cache replacement as an imitation learning problem to approximate the ``oracle" decisions of Belady's optimal policy.
These policies result in high implementation complexity and overhead, as they require online neural network inference.

Therefore, a second line of research uses powerful, unconstrained ML models in an offline setting to discover features and insights, which are used to design hardware-friendly online policies~\cite{shi19, sethumurugan2021designing}. For example, the Glider policy~\cite{shi19} leveraged a powerful, attention-based Long Short-Term Memory (LSTM) network to discover novel features for a simple online model.
In this case, the ML model serves as a ``discovery engine," but interpreting the model and designing the final simple policy remains a manual process.

Our work is the first to bridge the divide between these two schools.
We introduce an evolutionary coding framework that automates the discovery process of the ``Offline Insight" school, while explicitly optimizing for the simplicity and hardware-efficiency that is the key bottleneck for the ``Online Learning" school.

\subsubsection{Evolutionary Computation for Replacement Policy Tuning}

Prior work that has used evolutionary computation to discover replacement policies has focused on using genetic algorithms to tune parameters within a predefined policy structure~\cite{butt2010genetic, mourad2020novel, zadnik2010evolution}. While these works established a precedent for using evolutionary search, they were limited to optimizing parameters within a fixed-model, rather than evolving the structure of the algorithm itself.
More recently, an evolutionary coding framework was used to build web-based caching policies~\cite{dwivedula2025man}. This work uses an approach that is similar to ours, but applies it to software caches, where the constraints and performance tradeoffs are different.

\subsection{ML For Other Design Tasks}

Beyond predictive policies like cache replacement or branch prediction, ML has been used across a wide-variety of hardware-software co-design domains.

With the growing importance of ML accelerators, multiple prior works have focused on hardware-software co-design of neural network accelerators through various search space optimization techniques~\cite{sakhuja2023leveraging, murali20243dnn, xiao2021hasco, xiao2025core, 10.1145/3503222.3507767}.
These works often formulate the search problem to explore both hardware mixes and software mappings to newly generated hardware, leveraging advanced analytical modeling to quickly iterate through the design space.
Our work differs from these by focusing on {\em policies}, with a well defined set of championship constraints that resulting policies need to be evaluated with a detailed microarchitectural simulator running a mix of standard and realistic hyperscale CPU workloads.

Other techniques focused on optimizing microarchitectures~\cite{bai2024towards}, instead focus on other reinforcement learning techniques to explore pre-silicon parameters of microarchitectures or in the case of software, optimizing existing codebases/systems~\cite{cheng2025barbarians, massalin1987superoptimizer, 10.5555/645845.668618}.
Our work is different from these approaches as it tackles the whole-scale design (of a cache replacement policy) rather than parametric optimization of an existing system using LLM-based discovery tooling.

\section{Conclusions}
\label{sec:conclusion}

In this paper, we have introduced {\alphaarchitect}, a first step towards building an agentic artificial intelligence system for automatic novel computer architecture discovery.
Using {\alphaarchitect}, we automatically designed and implemented four novel cache replacement policy algorithms, {\policyone}, {\policyonept}, {\policythree}, {\policyfour}, that beat SoTA cache replacement policies tested in various cache replacement championships.
On the public multi-core Google Workload Traces, {\alphaarchitect} achieved a \num{5.322}\% better IPC speedup over the prior SoTA within two days, and on the single-core SPEC06 workloads, it achieved a \num{0.907}\% better IPC speedup over the prior SoTA in 18 days. Comparing against the effort involved in developing the prior SoTA policies, {\alphaarchitect} achieved these gains 3-5$\times$ faster than humans. We identified that agentic flows such as {\alphaarchitect} enable post-silicon hyperspecializtion of hardware systems to a specific workload (mix) demonstrating a  \num{2.374}\% IPC speedup improvement over prior SoTA on single-core SPEC06 workloads.
While we highlight that the use of these agentic systems allows for a dramatic increase in computer architecture discovery, further improvements are needed to improve simulator fidelity and speed in the era of agentic-AI-assisted research.

\section{Acknowledgments}

This paper has used generative AI technologies in the following ways.
First, the core work done by this paper uses AI tooling to develop novel cache replacement policies.
Minor use of generative AI tooling was used to generate tables, plot graphs, and assist with clarity and flow, limited to phrases and single sentences.
The authors have reviewed, verified, and edited all such generated content and take full responsibility for the content of the paper, including any errors or omissions.

\balance

\bibliographystyle{ACM-Reference-Format}
\bibliography{refs}

\end{document}